\documentclass[graybox,footinfo]{svmult}

\smartqed
\usepackage{mathptmx}       %
\usepackage{helvet}         %
\usepackage{courier}        %
\usepackage{type1cm}        %
\usepackage{graphicx}
\graphicspath{{./}{DahmKellerImages/}}
\usepackage{amsmath,amsfonts,amssymb,bm} %
\DeclareFontFamily{U}{mathx}{\hyphenchar\font45}
\DeclareFontShape{U}{mathx}{m}{n}{
      <5> <6> <7> <8> <9> <10>
      <10.95> <12> <14.4> <17.28> <20.74> <24.88>
      mathx10
      }{}
\DeclareSymbolFont{mathx}{U}{mathx}{m}{n}
\DeclareFontSubstitution{U}{mathx}{m}{n}
\DeclareMathAccent{\widecheck}      {0}{mathx}{"71}

\DeclareSymbolFont{bbold}{U}{bbold}{m}{n}
\DeclareSymbolFontAlphabet{\mathbbold}{bbold}

\usepackage{microtype} %

\usepackage{xcolor}
\definecolor{HighlightColor}{rgb}{0.69,0.81,0.11} %
\usepackage[pdftex,colorlinks,linkcolor=HighlightColor,citecolor=HighlightColor,filecolor=HighlightColor,urlcolor=HighlightColor,backref,pdfpagelabels,pdfpagemode=UseNone,implicit]{hyperref}
\usepackage{bookmark}
\makeatletter %
  \providecommand*{\toclevel@author}{999}
  \providecommand*{\toclevel@title}{0}
\makeatother

\usepackage[ruled,vlined]{algorithm2e}
\definecolor{shadecolor}{rgb}{0.43, 0.69, 0.26}

\usepackage{tikz}
\usetikzlibrary{calc,patterns,decorations.pathmorphing,decorations.markings,shapes,arrows}

\begin{document}

\title*{\textbf{Learning Light Transport the Reinforced Way}} %
\author{Ken Dahm \and Alexander Keller}
\institute{Ken Dahm \and Alexander Keller \at NVIDIA, Fasanenstr. 81, 10623 Berlin, Germany
\email{ken.dahm@gmail.com}, \email{keller.alexander@gmail.com}}

\maketitle

\abstract{We show that the equations of reinforcement learning
and light transport simulation are related integral equations. Based
on this correspondence, a scheme to learn importance while sampling
path space is derived. The new approach is demonstrated in a consistent
light transport simulation algorithm that uses reinforcement learning
to progressively learn where light comes from. As using this information
for importance sampling includes information about visibility, too,
the number of light transport paths with zero
contribution is dramatically reduced, resulting in much less noisy images
within a fixed time budget. }

\section{Introduction}

One application of light transport simulation is the computational
synthesis of images that cannot be distinguished from real photographs.
In such simulation algorithms \cite{PBRT}, light transport is modeled by a Fredholm
integral equation of the second kind and pixel colors are determined by
estimating functionals of the solution of the Fredholm integral equation.
The estimators are averages of contributions of sampled light transport paths
that connect light sources and camera sensors.

Compared to reality, where photons and their trajectories
are abundant, a computer may only consider a tiny fraction of path space,
which is one of the dominant reasons for noisy images. It is therefore
crucial to efficiently find light transport paths that have an important
contribution to the image.
While a lot of research in computer graphics has been focussing on
importance sampling
\cite{Lafortune95-FTRVM,PotentialDrivenMC,TableDrivenImpSampling,SignificanceCache,SequentialMCAdaptation},
for long there has not been a simple and efficient
online method that can substantially reduce the number of light
transport paths with zero contribution \cite{Vorba:2014:OLP}. %

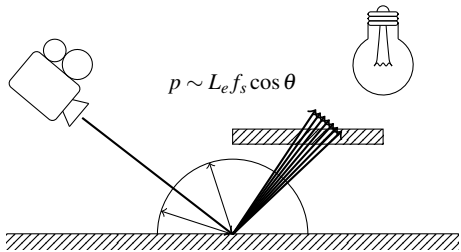
\begin{figure}[t]
  \centering
					\begin{tikzpicture}[scale=1]
							\node (wall1) [pattern = north east lines, anchor = center, minimum height = 0.2cm, minimum width = 6cm, rotate=0] at (0,-0.1cm) {};
							\draw (wall1.north east) -- (wall1.north west);

							\begin{scope}[shift={(1.8,2.6)},rotate=180,scale=0.4]
								\draw (0.0,0.0) arc (-60:240:1.0);
								\draw[rounded corners=1pt] (-1.0,-0.60) rectangle ++(1.0,0.15);
								\draw[rounded corners=1pt] (-1.0,-0.75) rectangle ++(1.0,0.15);
								\draw[rounded corners=1pt] (-1.0,-0.90) rectangle ++(1.0,0.15);
								\draw (-1.0,-0.5) -- (-1.0,0.0);
								\draw (0.0,-0.5) -- (0.0,0.0);
								\draw (-0.3,-0.9) arc (0:-180:0.2);
								\draw plot [smooth] coordinates {(-0.4,-0.45) (-0.3,0.675) (-0.2,0.9)};
								\draw plot [smooth] coordinates {(-0.6,-0.45) (-0.7,0.675) (-0.8,0.9)};
								\draw plot [smooth] coordinates {(-0.2,0.9) (-0.3,0.8) (-0.4,0.9) (-0.5,0.8) (-0.6,0.9) (-0.7,0.8) (-0.8,0.9)};
							\end{scope}

							\begin{scope}[shift={(-3,2)},rotate=-40]
								\draw[rounded corners=2pt] (0,0) rectangle ++(0.8,0.6);
								\draw (0.8,0.4) -- (1.0,0.5);
								\draw (0.8,0.2) -- (1.0,0.1);
								\draw (1.0,0.1) -- (1.0,0.5);
								\draw (0.2,0.75) circle (0.15);
								\draw (0.55,0.8) circle (0.2);
							\end{scope}

							\def\scatterRays{9}
							\draw (1.0,0.0) arc (0:180:1.0);
							\draw [->] (0,0) -- ({sin((1*180 - (\scatterRays+1)*90)/(\scatterRays+1)) * 1.0},{cos((1*180 - (\scatterRays+1)*90)/(\scatterRays+1)) * 1.0});
							\draw [->] (0,0) -- ({sin((4*180 - (\scatterRays+1)*90)/(\scatterRays+1)) * 1.0},{cos((4*180 - (\scatterRays+1)*90)/(\scatterRays+1)) * 1.0});
							
							\foreach \j in {1,...,7}{
								\draw [->, thick] (0,0) -- ({sin(((6.75+\j*0.12)*180 - (\scatterRays+1)*90)/(\scatterRays+1)) * 2.0},{cos(((6.75+\j*0.12)*180 - (\scatterRays+1)*90)/(\scatterRays+1)) * 2.0});
							}

							\draw [->,thick] (-2cm, 1.55cm) -- (0cm,0cm);
							\node at (0,2) {$p \sim L_e f_s\cos \theta$};

							\draw [pattern = north east lines] (0.0,1.2) rectangle ++(2.0,0.2);
					\end{tikzpicture}
  \caption{In the illustration, radiance
  is integrated by sampling proportional to the product of emitted radiance $L_e$
  and the bidirectional scattering distribution function $f_s$ representing the
  physical surface properties taking into account the fraction of radiance that
  is incident perpendicular to the surface, which is the cosine of the angle $\theta$
  between the surface normal and the direction of incidence. As such importance
  sampling does not consider blockers, light transport paths with zero contributions
  cannot be avoided unless visibility is considered.
  \label{Fig:Issue}}
\end{figure}

The majority of zero contributions are caused by unsuitable local
importance sampling using only a
factor instead of the complete integrand (see Fig.~\ref{Fig:Issue}) or
by trying to connect vertices of light transport path segments
that are occluded, for example shooting shadow rays to light
sources or connecting path segments starting both from the
light sources and the camera sensors. An example for this
inefficiency has been investigated early on in computer graphics
\cite{Veach95-OCSTM,Veach:1997:MLT}: The visible part of the synthetic scene shown in
Fig.~\ref{Fig:MaxVSAverage} is lit through a door. By closing
the door more and more the problem can be made arbitrarily
more difficult to solve.

We therefore propose a method that is based on reinforcement
learning \cite{ReinforcementLearning} and allows one to sample
light transport paths that are much more likely to connect lights
and sensors.
Complementary to first approaches of applying machine learning
to image synthesis \cite{Vorba:2014:OLP}, in Sec.~\ref{Sec:LightAndReinforcement} we show that %
light transport and reinforcement learning can be modeled by the
same integral equation. As a consequence, importance in
light transport can be learned using any light transport algorithm.

Deriving a relationship between reinforcement learning
and light transport simulation, we establish an automatic importance sampling
scheme as introduced in Sec.~\ref{Sec:Learning}.
Our approach allows for controlling the memory footprint, for suitable representations
of importance does not require
preprocessing, and can be applied during image synthesis and/or across
frames, because it is able to track distributions over time. A second parallel
between temporal difference learning and next event estimation is
pointed out in Sec.~\ref{Sec:NEE}.

As demonstrated in Sec.~\ref{Sec:Discussion} and shown in Fig.~\ref{Fig:Efficiency}, already a simple
implementation can dramatically improve light transport simulation.
The efficiency of the scheme is based on two facts: Instead of shooting towards the
light sources, we are guiding
light transport paths to where the light comes from, which effectively
shortens path length, and we learn importance from a smoothed
approximation instead from higher variance path space samples \cite{Lafortune95-FTRVM,Wan:95a,PracticalGuiding}.

\section{Identifying $Q$-Learning and Light Transport} \label{Sec:LightAndReinforcement}

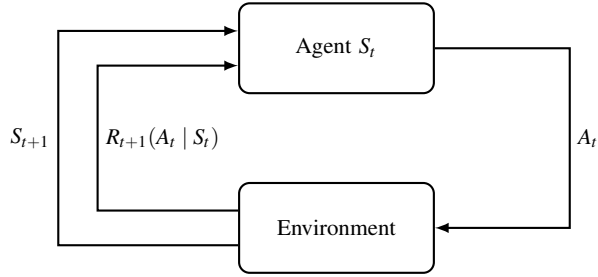
\begin{figure}[t]
    \centering
    \tikzstyle{block} = [rectangle, draw, text width=8em, text centered, rounded corners, minimum height=4em]
    \tikzstyle{line} = [draw, -latex]
    \begin{tikzpicture}[node distance = 8em, auto, thick]
    \node [block] (Agent) {Agent $S_t$};
    \node [block, below of=Agent] (Environment) {Environment};
     \path [line] (Agent.0) --++ (6em,0em) |- node [near start]{$A_t$} (Environment.0);
     \path [line] (Environment.190) --++ (-8em,0em) |- node [near start] { $S_{t+1}$} (Agent.170);
     \path [line] (Environment.170) --++ (-6.25em,0em) |- node [near start, right] {$R_{t+1}(A_t \mid S_t)$} (Agent.190);
    \end{tikzpicture}
    
    \caption{The setting for reinforcement learning: At time $t$, an agent is in state $S_t$
    and takes an action $A_t$, which after interaction with the environment
    brings him to the next state $S_{t + 1}$ with a scalar reward $R_{t+1}$.
    \label{Fig:RL}}
\end{figure}

The setting of reinforcement learning \cite{ReinforcementLearning}
is depicted in Fig.~\ref{Fig:RL}:
An agent takes an action thereby transitioning to the resulting
next state and receiving a reward. In order to maximize the reward,
the agent has to learn which action to choose in what state.
This process very much resembles how humans learn.

$Q$-learning \cite{Qlearning} is a model free reinforcement learning technique.
Given a set of states $S$ and a set of actions $A$, 
it determines
a function $Q(s, a)$ that for any $s \in S$ values taking
the action $a \in A$. Thus given a state $s$, the action $a$ with the highest
value may be selected next and
\begin{equation} \label{Eqn:Qmax}
  Q(s, a) = (1 - \alpha) \cdot Q(s, a) + \alpha \cdot \left(r(s, a) + \gamma \cdot \max_{a' \in A} Q(s', a') \right)
\end{equation}
may be updated by a fraction of $\alpha \in [0,1]$, where $r(s, a)$ is the reward
for taking the action resulting in a transition to a state $s'$. In addition,
the maximum $Q$-value of possible actions in $s'$ is considered and discounted
by a factor of $\gamma \in [0,1)$. %

\begin{figure}
    \centering
    \begin{tabular}{cc}
    \includegraphics[width=0.495\linewidth]{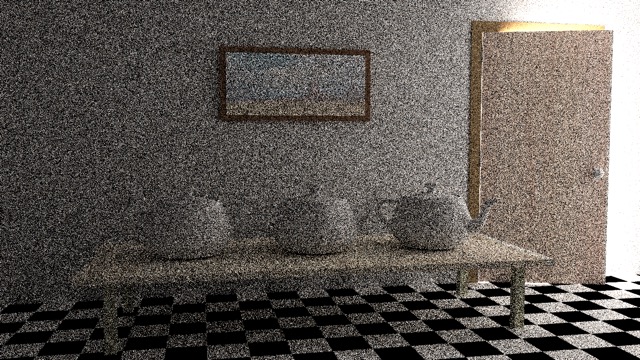}
    &
    \includegraphics[width=0.495\linewidth]{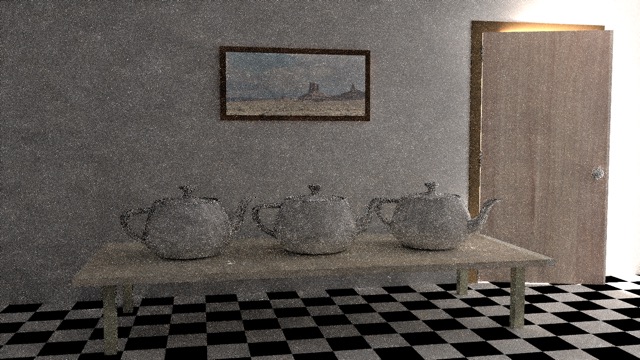}
    \\ a & b \\
    \includegraphics[width=0.495\linewidth]{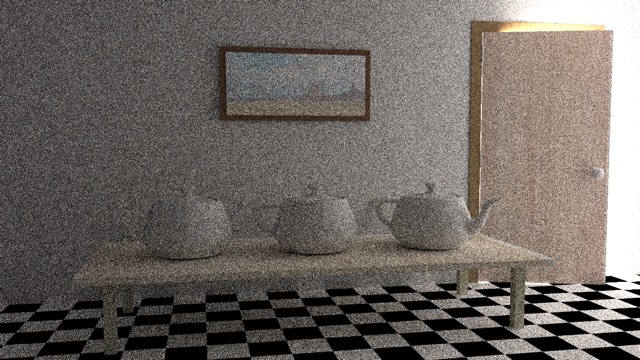}
    &
    \includegraphics[width=0.495\linewidth]{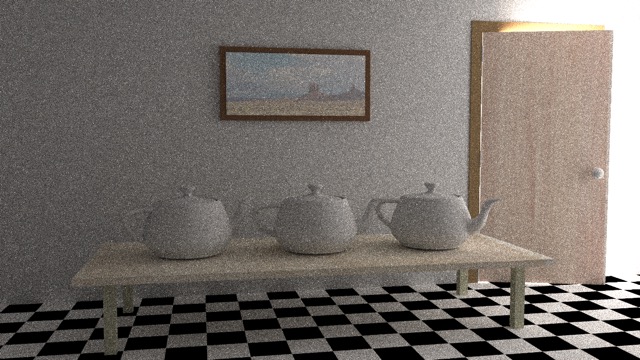}
    \\ c & d \\
    \includegraphics[width=0.495\linewidth]{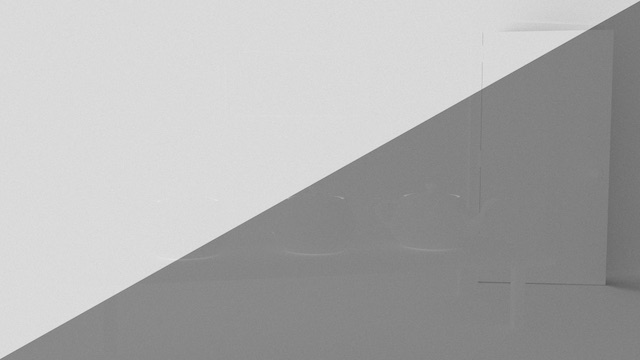}
    &
    \includegraphics[width=0.495\linewidth]{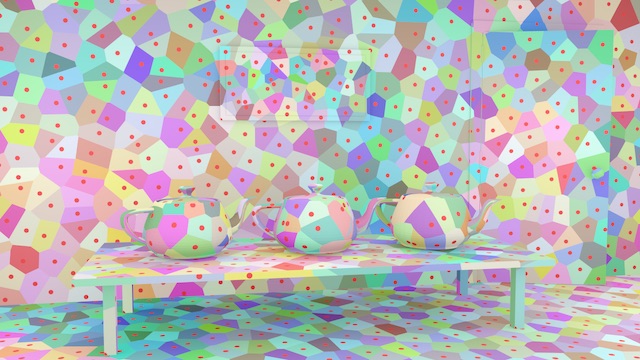}
    \\ e & f
    \end{tabular}
    
    \caption{Comparison at 1024 paths per pixel (the room behind the
    door is shown in Fig.~\ref{Fig:Distribution}):
    a) A simple path tracer with cosine importance sampling,
    b) the Kelemen variant of the Metropolis light transport algorithm,
    c) scattering proportional to $Q$, while
    updating $Q$ with the maximum as in Eqn.~\ref{Eqn:Qmax} and
    d) scattering proportional to $Q$ weighted by the bidirectional
    scattering distribution function and updating accordingly by Eqn.~\ref{Eqn:Learning}. The predominant reinforcement
    approach of always taking the best next action is inferior to selecting the next action
    proportional to $Q$, i.e. considering all alternatives. A comparison to the Metropolis
    algorithm reveals much more uniform lighting, especially much more uniform noise and the lack of the typical splotches.
    e) The average path length of path tracing (above image diagonal) is about 215, while
    with reinforcement learning it amounts to an average of 134. The average path
    length thus is reduced by 40\% in this scene.
    f) Discretized hemispheres to approximate $Q$ are stored in points on the scene surfaces
    determined by samples of the Hammersley low discrepancy
    point set. Retrieving $Q$ for a query point results in searching
    for the nearest sample of $Q$ that has a similar normal to
    the one in the query point (see especially the teapot handles). The red points indicate where in the
    scene hemispheres to hold the $Q_k$ are stored. The colored areas indicate
    their corresponding Voronoi cells. Storing the $Q_k$ in this example
    requires about 2 MBytes of memory. Scene courtesy
    (cc) 2013 Miika Aittala, Samuli Laine, and Jaakko Lehtinen (https://mediatech.aalto.fi/publications/graphics/GMLT/).
    \label{Fig:MaxVSAverage}}
\end{figure}

Instead of taking into account only the best valued action,
\[
  Q(s, a) = (1 - \alpha) \cdot Q(s, a) + \alpha \cdot \left(r(s, a) + \gamma \cdot \sum_{a' \in A} \pi(s', a') Q(s', a') \right)
\]
averages all possible actions in $s'$ and weighs their values $Q(s', a')$ by a
transition kernel $\pi(s', a')$, which is a strategy called {\em expected SARSA} \cite[Sec.6.6]{ReinforcementLearning}. This is especially interesting, as later it will turn
out that always selecting the "best" action does not perform as well as considering
all options (see Fig.~\ref{Fig:MaxVSAverage}).
For a continuous space $A$ of actions, we then have
\begin{equation} \label{Eqn:Continuous}
  Q(s, a) = (1 - \alpha) \cdot Q(s, a) + \alpha \cdot \left(r(s, a) + \gamma \cdot \int_{A} \pi(s', a') Q(s', a') da' \right) .
\end{equation}
On the other hand, the radiance
\begin{equation} \label{Eqn:Radiance}
  L(x, \omega) = L_e(x, \omega) + \int_{{\mathcal S}^+(x)} L(h(x, \omega_i), -\omega_i) f_s(\omega_i, x, \omega) \cos \theta_i d\omega_i
\end{equation}
in a point $x$ on a surface into direction $\omega$ is modeled by a Fredholm integral equation
of the second kind. $L_e$ is the source radiance and the integral
accounts for all radiance that is incident over the hemisphere ${\mathcal S}^+(x)$
aligned by the surface normal in $x$ and transported into direction $\omega$.
The hitpoint function $h(x, \omega_i)$ traces a ray from $x$ into direction $\omega_i$
and returns the first surface point intersected. The radiance from this point
is attenuated by the bidirectional scattering distribution function $f_s$,
where the cosine term of the angle $\theta_i$ between surface normal and
$\omega_i$  accounts for only the fraction that is perpendicular to the
surface.

A comparison of Eqn.~\ref{Eqn:Continuous} for $\alpha = 1$ and
Eqn.~\ref{Eqn:Radiance} reveals structural similarities of the formulation
of reinforcement learning and the light transport integral equation,
respectively, which lend themselves to matching terms:
Interpreting the state $s$ as a location $x \in \mathbb{R}^3$ and an action $a$
as tracing a ray from location $x$ into direction $\omega$ resulting in the point
$y := h(x, \omega)$ corresponding to the state $s'$, the reward term $r(s, a)$
can be linked to the emitted radiance $L_e(y, -\omega) = L_e(h(x, \omega), -\omega)$
as observed from $x$. Similarly, the integral operator can be applied to
the value $Q$, yielding
\begin{equation} \label{Eqn:Q}
  Q(x, \omega) =  L_e(y, -\omega) + \int_{{\mathcal S}^+(y)} Q(y, \omega_i ) f_s(\omega_i, y, -\omega) \cos \theta_i  d\omega_i ,
\end{equation}
where we identified the discount factor $\gamma$ multiplied by the policy $\pi$ and
the bidirectional scattering distribution function $f_s$.
Taking a look at the geometry and the physical meaning of the terms, it becomes
obvious that $Q$ in fact must be the radiance $L_i(x, \omega)$ incident
in $x$ from direction $\omega$ and in fact is described by a Fredholm integral
equation of the second kind - like the light transport equation~\ref{Eqn:Radiance}.
\begin{figure}
    \centering

	\includegraphics[width=0.497\linewidth]{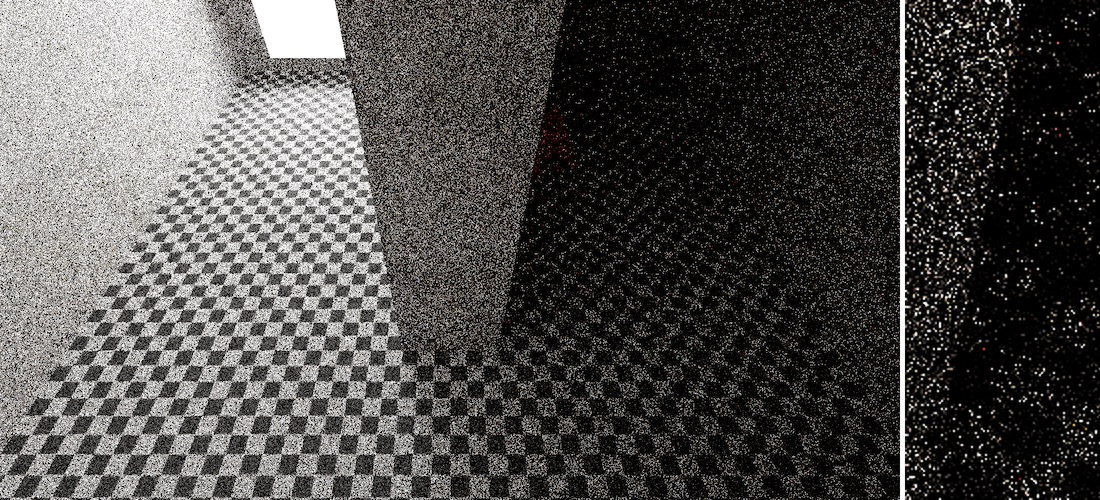} \hfill
    \includegraphics[width=0.497\linewidth]{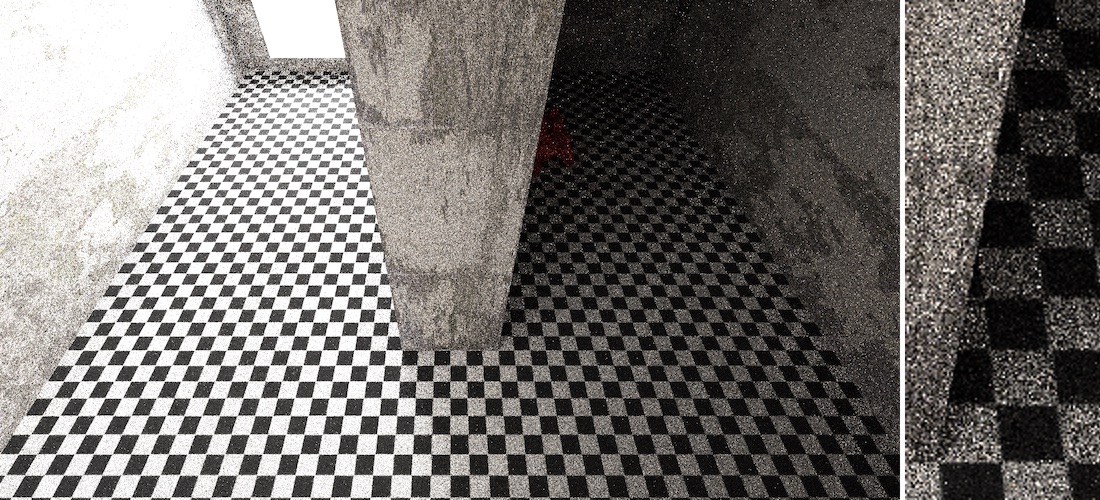}

	\includegraphics[width=0.497\linewidth]{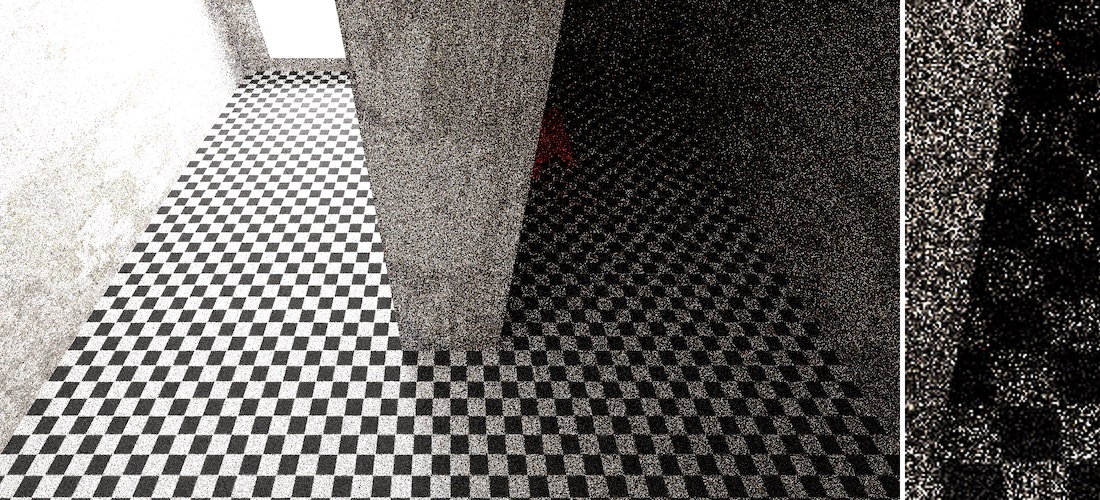} %
    \hfill
	\includegraphics[width=0.497\linewidth]{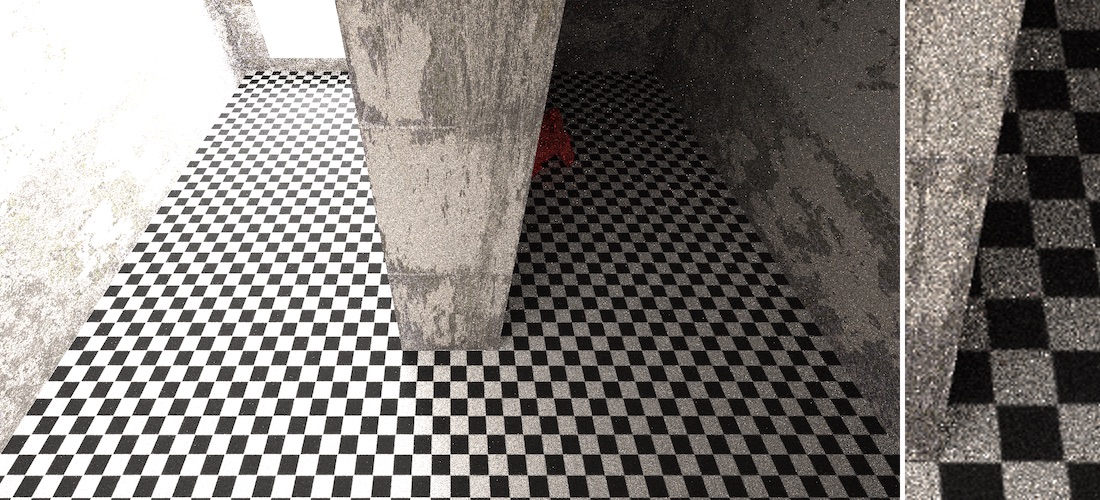} %
    
    \caption{Comparison of simple path tracing without (left) and with (right)
    reinforcement learning importance sampling. The top row is using 8 paths
    per pixel, while 32 are used for the bottom row. The challenge of the scene
    is the area light source on the left indirectly illuminating the right part of the scene.
    The enlarged insets illustrate the reduction of noise level.
    \label{Fig:ShowOff}}
\end{figure}

\section{$Q$-Learning while Path Tracing} \label{Sec:Learning}

In order to synthesize images, we need to compute functionals
of the radiance equation \ref{Eqn:Radiance},
i.e. project the radiance onto the image plane. For the purpose of this article,
we start with a simple forward path tracer \cite{PBRT,NutshellQMC}: From a virtual camera,
rays are traced through the pixels of the screen. Upon their first
intersection with the scene geometry,  the light transport path is
continued into a scattering direction determined according to the optical surface
properties. Scattering and ray tracing are repeated until a light source
is hit. The contribution of this complete light transport path is added to
the pixel pierced by the initial ray of this light transport path when started at the camera.

In this simple form, the algorithm exposes quite some variance as
can be seen in the images on the left in Fig.~\ref{Fig:ShowOff}. This noise
may be reduced by importance sampling. We therefore progressively approximate
Eqn.~\ref{Eqn:Q} using reinforcement learning: Once a direction has been
selected and a ray has been traced by the path tracer,
\begin{eqnarray}
  Q'(x, \omega) & = & (1 - \alpha) \cdot Q(x, \omega) \label{Eqn:Learning} \\
  && + \alpha \cdot \left(L_e(y, -\omega) + \int_{{\mathcal S}^+(y)} Q(y, \omega_i ) f_s(\omega_i, y, -\omega) \cos \theta_i  d\omega_i\right) \nonumber
\end{eqnarray}
is updated using a learning rate $\alpha$. The probability density function
resulting from normalizing $Q$ in turn is used for importance sampling
a direction to continue the path. As a consequence more and more light
transport paths are sampled that contribute to the image. Computing
a global solution to $Q$ in a preprocess would not allow for focussing
computations on light transport paths that contribute to the image.

\begin{figure}[t]
    \centering
    \includegraphics[width=\linewidth]{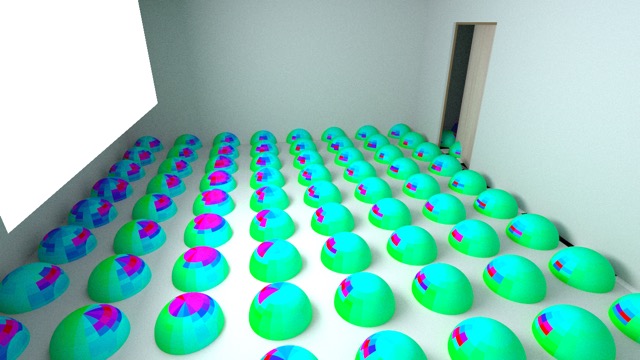}
        
    \caption{The image shows parts of an example discretization of $Q$ by a grid, %
    where hemispheres are uniformly distributed across the ground plane. The
    false colors indicate magnitude, where small values are green and
    large values are red. The large values on each hemisphere
    point towards the part of the scene, where the light is coming from.
    For example, under the big area light source on the left, most radiance is
    incident as reflected radiance from the wall opposite to the light source.
    \label{Fig:Distribution}}
\end{figure}

\begin{algorithm}[t]
							\DontPrintSemicolon
							\footnotesize
							\SetKwProg{Fn}{Function}{}{}
							\SetKwProg{ElseIf}{else if}{}{}
							\Fn{pathTrace($camera, scene$)}{
								$throughput$ $\gets$ 1\;
								$ray$ $\gets$ setupPrimaryRay($camera$)\;
								\For{$i\leftarrow 0$ \KwTo $\infty$}{
									$(y, n)$ $\gets$ intersect($scene$, $ray$)\tcp{$y := h(x, \omega)$}
									{%
									\tcp{addition 1: update $Q$}
									\If{$i > 0$}{
						$
							Q'(x, \omega) = (1 - \alpha) Q(x, \omega) + \alpha \left(L_e(y, -\omega) + \int_{{\mathcal S}^2_+(y)} f_s(\omega_i, y, -\omega) \cos \theta_i Q(y, \omega_i ) d\omega_i\right)
						$\;
										}}
									
									\If{isEnvironment($y$)}{
										return $throughput \cdot$ getRadianceFromEnvironment($ray, y$)\;
									}
									\ElseIf{isAreaLight($y$)}{
										return $throughput \cdot$ getRadianceFromAreaLight($ray, y$)\;
									}
									\tcp{addition 2: scatter proportional to $Q$}
									{$(\omega, p_{\omega}, f_s) \gets$ {\textbf{%
									sampleScatteringDirectionProportionalToQ($y$)}} \;}
									$throughput$ $\gets$ $throughput \cdot f_s \cdot \cos(n, \omega)$ / $p_{\omega}$\;
									$ray \gets (y, \omega)$\;
								}
							}
							    
    \caption{Augmenting a path tracer by reinforcement learning
    for importance sampling requires only two additions:
    The importance $Q$ needs to be updated along the path
    and scattering directions are sampled proportional to $Q$ as
    learned so far.\label{Alg:Pathtracer}}
\end{algorithm}

\subsection{Implementation} \label{Sec:Implementation}

Often, approximations to $Q$ are tabulated for each pair of
state and action. In computer graphics, there are multiple choices
to represent radiance and for the purpose of this article, we chose
the data structure as used for irradiance volumes
\cite{IrradianceVolume} to approximate $Q$.
Fig.~\ref{Fig:Distribution} shows an exemplary visualization of such a discretization
during rendering: For selected points $y$ in space, the hemisphere is stratified and
one value $Q_k(y)$ is stored per sector, i.e. stratum $k$.
Fig.~\ref{Fig:MaxVSAverage}f illustrates the placement of probe centers $y$,
which results from  mapping a two-dimensional low discrepancy sequence onto the scene
surface. %

Now the integral 
\begin{eqnarray*}
  \int_{{\mathcal S}^+(y)} Q(y, \omega_i) f_s(\omega_i, y, -\omega) \cos \theta_i  d\omega_i
  \approx \frac{2 \pi}{n} \sum_{k = 0}^{n - 1} Q_k(y) f_s(\omega_k, y, -\omega) \cos \theta_k
\end{eqnarray*}
in Eqn.~\ref{Eqn:Learning} can be estimated
by using each one uniform random direction $\omega_k$ in each
stratum $k$, where $\theta_k$ is the angle between the surface normal in $y$
and $\omega_k$.

The method has been implemented in an importance driven forward path tracer
as shown in Alg.~\ref{Alg:Pathtracer}: Only two routines for updating
$Q$ and selecting a scattering direction proportional to $Q$ need to be added.
Normalizing the $Q$ in a point $y$ then results in a probability density
that is used for importance sampling during scattering by inverting the cumulative
distribution function. In order to guarantee ergodicity, meaning that every light transport
path remains possible, all $Q(y)$ are initialized to be positive, for example by a uniform
probability density or proportional to a factor of the integrand (see Fig.~\ref{Fig:Issue}).
When building the cumulative distribution functions in parallel every accumulated frame,
values below a small positive threshold are replaced by the threshold.

The parameters exposed by our implementation are the
resolution of the discretization and the learning rate $\alpha$.

\subsection{Consistency} %

It is desirable to craft consistent rendering algorithms \cite{NutshellQMC},
because then all renderer introduced artifacts, like for example noise,
are guaranteed to vanish over time. This requires the $Q_k(y)$ to converge,
which may be accomplished by a vanishing learning rate $\alpha$.

In reinforcement learning \cite{ReinforcementLearning}, a typical approach is to count the number
of visits to each pair of state $s$ and action $a$ and using
\[
  \alpha(s, a) = \frac{1}{1 + \text{visits}(s, a)} .
\]
The method resembles the one used to make progressive
photon mapping consistent \cite{Hachisuka:2008:PPM}, where consistency
has been achieved by decreasing the search radius around a
query point every time a photon hits sufficiently close. Similarly,
the learning rate may also depend on the total number of
visits to a state $s$ alone, or even may be chosen to vanish
independently of state and action. Again, such approaches have
been explored in consistent photon mapping \cite{OneSequence}.

While the $Q_k(y)$ converge, they do not necessarily converge to the
incident radiance in Eqn.~\ref{Eqn:Q}. First, as they are projections onto
a basis, the $Q_k(y)$ at best only are an approximation of $Q$ in realistic settings. Second, 
as the coefficients $Q_k(y)$ are learned during path tracing, i.e.
image synthesis, and used for importance sampling, it may well happen that they
are not updated everywhere at the same rate.
Nevertheless, since all operators are linear, the number of
visits will be proportional to the number of light transport paths \cite{OneSequence}
and consequently as long as $Q_k(y) > 0$ whenever $L_i(y, \omega_i) > 0$
all $Q_k(y)$ will be updated eventually.

\subsection{Learning while Light Tracing} \label{Sec:LightTracing} %

For guiding light transport paths from the light sources towards the camera,
the transported weight $W$ of a measurement (see \cite{Veach:1994:BELT}), i.e.
the characteristic function of the image plane,
has to be learned instead of the incident radiance $Q$.
As $W$ is the adjoint of $Q$, the same data structures may be used
for its storage.
Learning both $Q$ and
$W$ allows one to implement bidirectional path tracing \cite{Veach:1994:BELT}
with reinforcement learning for importance sampling to
guide both light and camera path segments including visibility information
for the first time. Note that guiding light transport paths this way
may reach efficiency levels that even can make bidirectional path tracing and multiple
importance sampling redundant
\cite{Vorba:2014:OLP} in many common cases.

\begin{figure}
    \centering
    \includegraphics[width=0.49\linewidth]{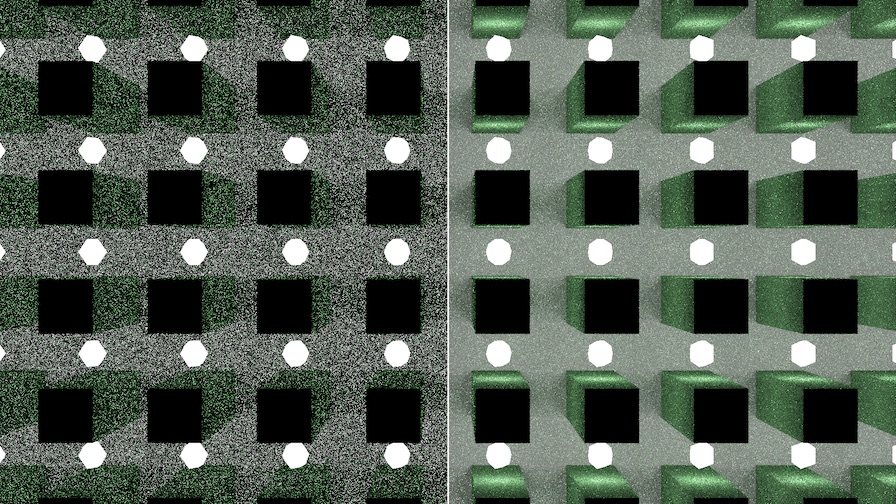} \hfill
    \includegraphics[width=0.49\linewidth]{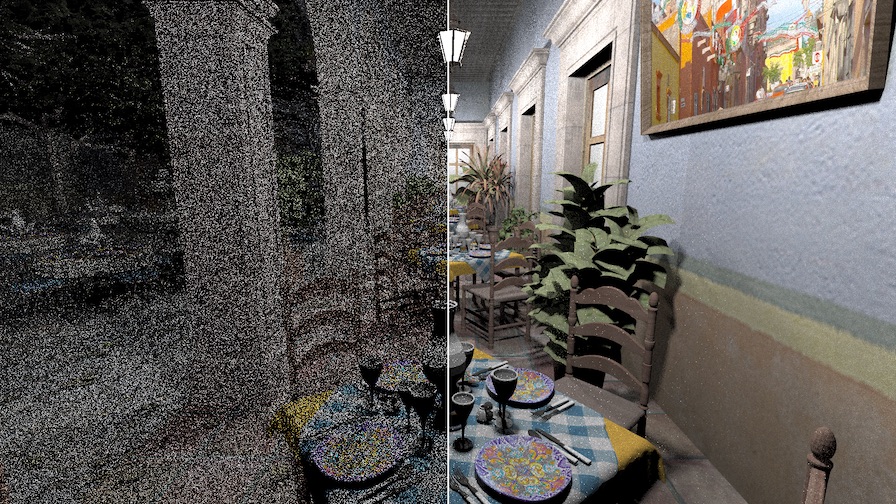}
    
    \caption{Two split-image comparisons of uniformly selecting area light
    sources and selection using temporal difference learning, both at 16 paths per pixel.
    The scene on the left has 5000 area light sources, whereas the scene
    on the right has about 15000 (San Miguel scene courtesy
    Guillermo M. Leal Llaguno (http://www.evvisual.com/)).
    \label{Fig:TD-NEE}}
\end{figure}

\section{Temporal Difference Learning and Next Event Estimation} \label{Sec:NEE} %

Besides the known shortcomings of (bidirectional) path tracing
\cite[Sec.2.4 Problem of insufficient techniques]{KK:01},
the efficiency may be restricted
by the approximation quality of $Q$: For example, the smaller the
light sources, the finer the required resolution of $Q$ to reliably guide
rays to hit a light source. This is where next event estimation
may help \cite{AdaptiveDirectIllumination,Iray_report,AdaptiveTreeSplitting}. %

Already in \cite{War:91} the contribution of light sources has
been ``learned'': A probability per light source has been
determined by the number of successful shadow
rays divided by the total number of shadow rays shot.
This idea has been refined subsequently \cite{Keller:00:Imp,IGI1,IGI2,WALDEGSR2003}.

For reinforcement learning, the state space may be chosen as
a regular grid over the scene, where in each grid cell $c$
for each light source $l$ a value $V_{c, l}$ is stored that
is initialized with zero. %
Whenever a sample on a light source $l$ is visible to a point $x$
to be illuminated in the cell $c$ upon next event estimation, its value
\begin{equation} \label{Eqn:TD}
  V_{c,l}' = (1-\alpha) V_{c,l} + \alpha \cdot \|C_l(x)\|_{\infty}%
\end{equation}
is updated using the norm of the contribution $C_l(x)$.
Building a cumulative distribution function from all
values $V_{c,l}$ within a cell $c$, light may be selected by importance sampling.
Fig.~\ref{Fig:TD-NEE} shows the efficiency gain of this
reinforcement learning method over uniform light source selection for 16 paths per pixel.

It is interesting to see that this is another relation to reinforcement
learning: While the $Q$-learning equation~\ref{Eqn:Learning}
takes into account the values of the next, non-terminal state, the next
state in event estimation is always a terminal state and
$Q$-learning coincides with plain temporal difference learning \cite{TDlearning}
as in equation~\ref{Eqn:TD}.

\subsection{Learning Virtual Point Light Sources}

The vertices generated by tracing photon trajectories
(see Sec.~\ref{Sec:LightTracing}) can be considered a photon map
\cite{PhotonMap:01} and may be used in the same way. Furthermore,
they may be used as a set
of virtual point light sources for example the instant radiosity
\cite{Keller97-IRad} algorithm. 

Continuously updating and learning the measurement
contribution function $W$ \cite{Veach:1994:BELT} across frames and
using the same seed for the pseudo- or quasi-random
sequences allows for generating virtual point light sources
that expose a certain coherency over time, which reduces
temporal artifacts when rendering animations with global illumination.

\begin{figure}
    \centering
    
    \includegraphics[height=0.397\linewidth]{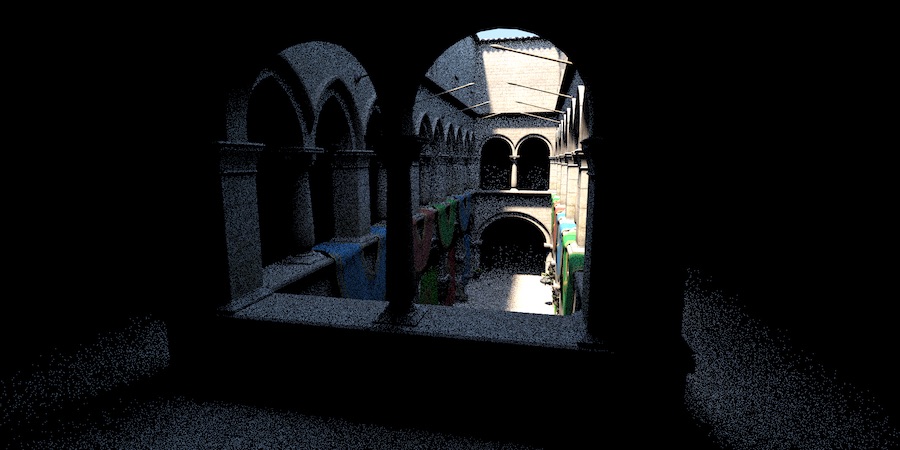} %
    \hfill \includegraphics[height=0.397\linewidth]{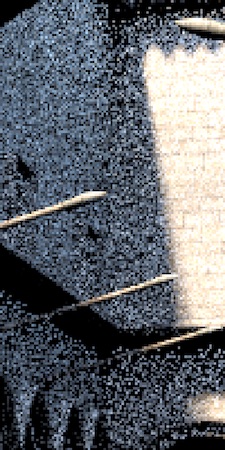}
    
    \includegraphics[height=0.397\linewidth]{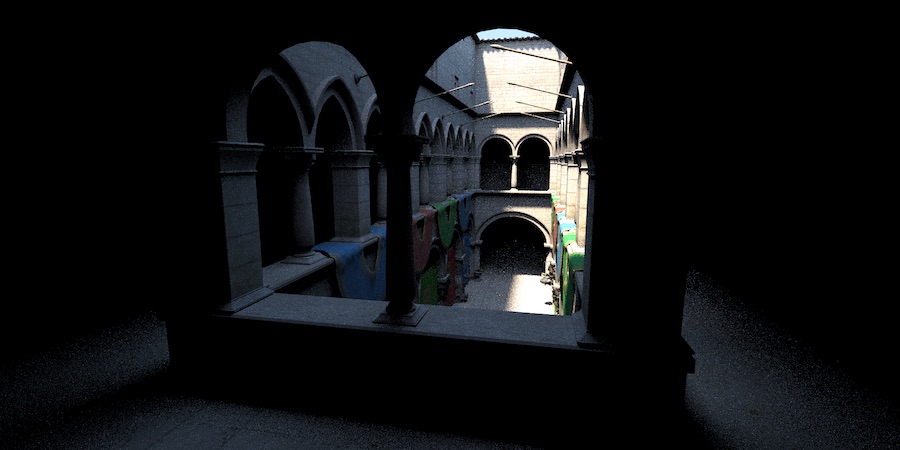} %
    \hfill \includegraphics[height=0.397\linewidth]{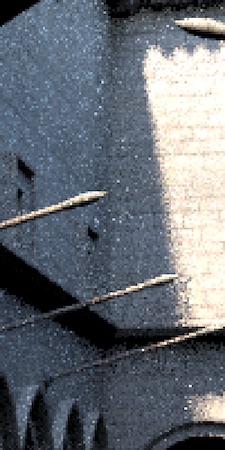}
    
    \caption{Sun and sky illumination at 32 paths per pixel.
    Top: simple importance sampling considering only the environment map as a light source. Bottom: Importance sampling
    with reinforcement learned importance. The enlargements on the right
    illustrate the improved noise reduction. Scene courtesy
    Frank Meinl, Crytek (http://graphics.cs.williams.edu/data/meshes/crytek-sponza-copyright.html).
    \label{Fig:Sunsky2}}
\end{figure}

\subsection{Learning Environment Lighting}

Rendering sun and sky is usually done by distributing samples
proportional to the brightness of pixels in the environment texture.
More samples should end up in brighter regions, which is achieved
by constructing and sampling from a cumulative distribution function,
for example using the alias method \cite{Alias91}. Furthermore, the sun may be
separated from the sky and simulated separately.
The efficiency of such importance sampling is highly dependent on occlusion,
i.e. what part of the environment can be seen from the point to be shaded
(see Fig.~\ref{Fig:Issue}).

Similar to Sec.~\ref{Sec:Implementation} and in order to consider the
actual contribution including occlusion, an action space is defined by
partitioning the environment map into tiles and learning the
importance per tile.
Fig.~\ref{Fig:Sunsky2} shows the improvement for an example setting.

\begin{figure}[t]
    \centering
    \includegraphics[width=\linewidth]{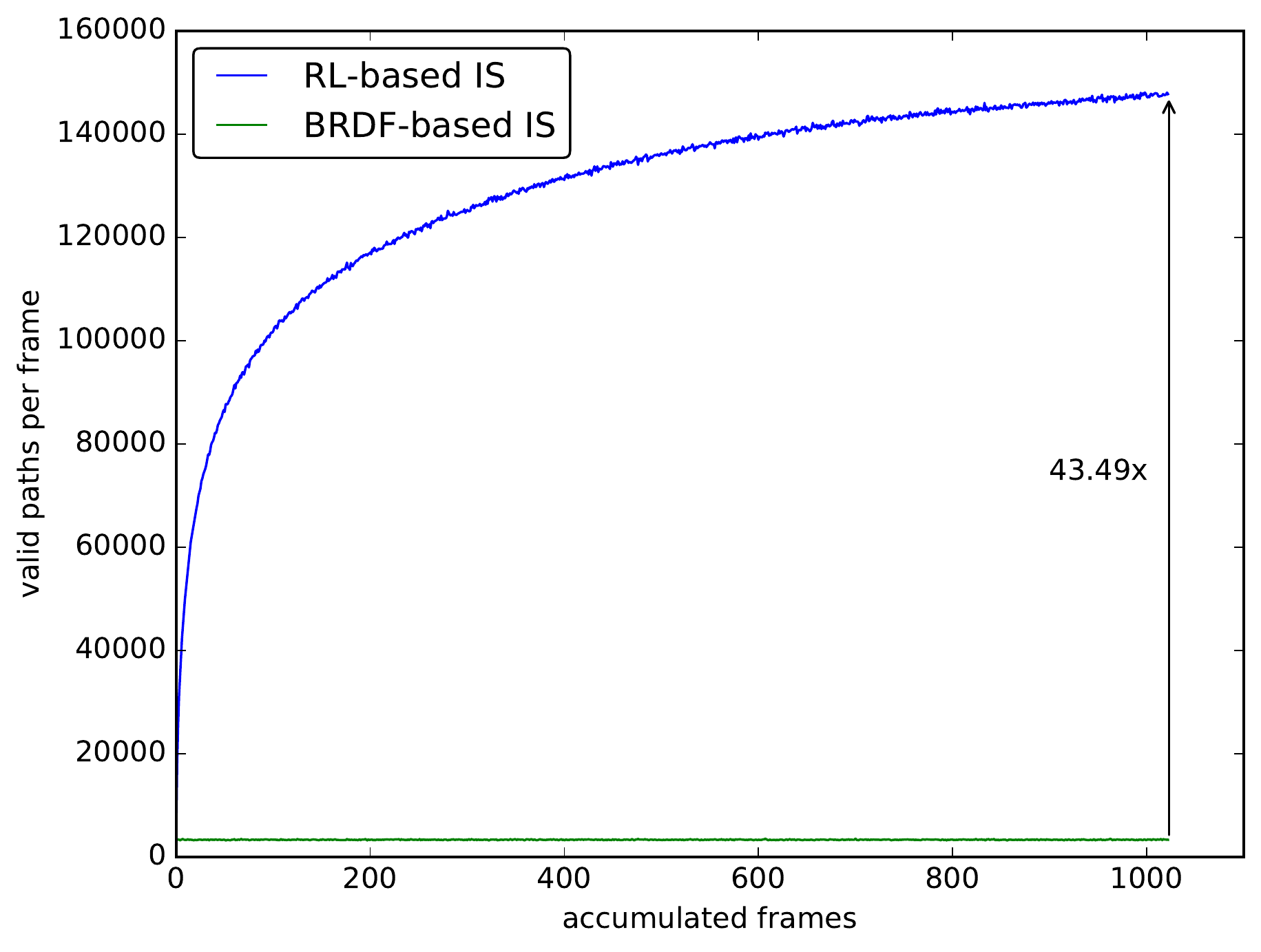}
    
    \caption{Using reinforcement learning (RL), the number of paths
    actually connecting to a light source is dramatically improved
    over classic importance sampling (IS) using only the bidirectional scattering
    distribution function for importance sampling.
    As a result, more non-zero contributions are accumulated
    for the same number of paths, see also Fig.~\ref{Fig:MaxVSAverage}.
    \label{Fig:Efficiency}}
\end{figure}

\section{Results and Discussion} \label{Sec:Discussion}

Fig.~\ref{Fig:MaxVSAverage} compares the new reinforcement
learning algorithm to common algorithms: For the same budget
of light transport paths, the superiority over path tracing with
importance sampling according to the reflection properties
is obvious. A comparison with the Metropolis algorithm for importance
sampling \cite{Veach:1997:MLT,KelemenMLT} reveals much
more uniform noise lacking the typical splotchy structure inherent
with the local space exploration of Metropolis samplers.
Note, however, that the new reinforcement learning importance
sampling scheme could as well be combined with Metropolis sampling.
Finally, updating $Q$ by Eqn.~\ref{Eqn:Qmax}, i.e. the ``best possible action'' strategy
is inferior to using the weighted average of all possible next actions
according to Eqn.~\ref{Eqn:Learning}. In light transport simulation this is
not surprising, as the deviation of the integrand from its estimated
maximum very often is much larger than from a piecewise constant
approximation.
The big gain in quality is due to the dramatic reduction of zero
contribution light transport paths (see Fig.~\ref{Fig:Efficiency}),
even under complex lighting. In
Figs.~\ref{Fig:MaxVSAverage}a-d, the same number of paths
has been used. In each iteration,
for path tracing with and without reinforcement learning one path
has been started per pixel, while for the Metropolis variant the number
of Markov chains equals the number of pixels of the image. Rendering
the image at 1280x720 pixels, each iteration takes
41ms for path tracing, 49ms for Metropolis light transport \cite{Veach:1997:MLT,KelemenMLT},
and 51ms for the algorithm with reinforcement learned importance
sampling. Hence the 20\% overhead is well paid off by the level of noise reduction.

Shooting towards where the radiance comes from naturally
shortens the average path length as can be seen in
Fig.~\ref{Fig:MaxVSAverage}e. 
Based on the approach to guide light paths using a
pre-trained Gaussian mixture model \cite{Vorba:2014:OLP}
to represent probabilities, in \cite{Vorba:2015:ADR}
in addition the density of light transport paths
is controlled across the scene using splitting and Russian
roulette. These ideas have the potential to further
improve the efficiency of our approach.

While the memory requirements for storing our data structure
for $Q$ are small, the data structure is not adaptive. An
alternative is an adaptive hierarchical approximation to $Q$
as used in \cite{Lafortune95-FTRVM,PracticalGuiding}. Yet, another variant
would be learning parameters for lobes to guide
light transport paths \cite{SignificanceCache}. In principle
any data structure that has been used in graphics to
approximate irradiance or radiance is a candidate.
Which data structure and what parameters are best, may
depend on the scene to be rendered. For example, using discretized
hemispheres limits the resolution with respect to
solid angle. If the resolution is chosen too fine, learning
is slow, if the resolution is to coarse, convergence is
slow.

Given that $Q$ asymptotically approximates the incident radiance $L_i$,
it is worthwhile to investigate how it can be used for the
separation of the main part as explored in \cite{Lafortune95-FTRVM,SequentialMCAdaptation}
to further speed up light transport simulation or even as an
alternative to importance sampling.

Beyond what we explore, path guiding has been
extended to consider product importance sampling \cite{PISGuiding} and
reinforcement learning \cite{ReinforcementLearning} offers
more policy evaluation strategies to consider.

\section{Conclusion}

Guiding light transport paths has been explored
in \cite{Lafortune95-FTRVM,Wan:95a,PotentialDrivenMC,SequentialMCAdaptation,SignificanceCache,Vorba:2014:OLP,PracticalGuiding}. 
However, key to our approach is that by using a representation
of $Q$ in Eqn.~\ref{Eqn:Learning} instead of solving the
equation by recursion, i.e. a Neumann series, $Q$ may be learned
much faster and in fact during sampling light transport paths
without any preprocess. 
This results in a new algorithm to increase the efficiency of
path tracing by approximating importance using reinforcement
learning during image synthesis. Identifying $Q$-learning and
light transport, heuristics have been replaced by physically
based functions, and the only parameters that the user may
control are the learning rate and the discretization of $Q$.

The combination of reinforcement learning and deep neural networks
\cite{AtariDeepRL,DoubleQ,ContinuousDRL,AsynchDeepRL} is an
obvious avenue of future research: Representing the
radiance on hemispheres already has been successfully
explored \cite{SkyANN} and the interesting question
is how well $Q$ can be represented by neural networks.

\begin{acknowledgement}
The authors would like to thank Jaroslav K\v{r}iv\'anek, Tero Karras, Toshiya Hachisuka,
and Adrien Gruson for profound discussions and comments.
\end{acknowledgement}

\end{document}